\documentclass[10pt,twocolumn,letterpaper]{article}

\usepackage{cvpr}              

\usepackage{graphicx}
\usepackage{amsmath}
\usepackage{amssymb}
\usepackage{booktabs}
\usepackage{diagbox}
\usepackage{multirow}
\usepackage{amsthm,amsmath,amssymb}
\usepackage{mathrsfs}

\usepackage[pagebackref,breaklinks,colorlinks]{hyperref}

\usepackage[capitalize]{cleveref}
\crefname{section}{Sec.}{Secs.}
\Crefname{section}{Section}{Sections}
\Crefname{table}{Table}{Tables}
\crefname{table}{Tab.}{Tabs.}

\def\cvprPaperID{1951} 
\def\confName{CVPR}
\def\confYear{2023}

\begin{document}

\title{\LaTeX\ Author Guidelines for \confName~Proceedings}

\title{Multi-view Adversarial Discriminator: Mine the Non-causal Factors for Object Detection in Unseen Domains}

\author{
    Mingjun Xu, Lingyun Qin, Weijie Chen, Shiliang Pu, Lei Zhang\thanks{Corresponding author (Lei Zhang)}\\
    School of Microelectronics and Communication Engineering, Chongqing University, China\\
    Hikvision Research Institute, Hangzhou, China\\
    {\tt\small \{mingjunxu, lingyunQin\}@cqu.edu.cn, \{chenweijie5, pushiliang.hri\}@hikvision.com,} \\
    {\tt\small leizhang@cqu.edu.cn}
}
\maketitle

\begin{abstract}
    Domain shift degrades the performance of object detection models in practical applications. 
    To alleviate the influence of domain shift, plenty of previous work try to decouple and learn the domain-invariant (common) features from source domains via domain adversarial learning (DAL).
    However, inspired by causal mechanisms, we find that previous methods ignore the implicit insignificant non-causal factors hidden in the common features. This is mainly due to the single-view nature of DAL. 
    In this work, we present an idea to remove non-causal factors from common features by multi-view adversarial training on source domains, because we observe that such insignificant non-causal factors may still be significant in other latent spaces (views) due to the multi-mode structure of data.
    To summarize, we propose a Multi-view Adversarial Discriminator (MAD) based domain generalization model, consisting of a Spurious Correlations Generator (SCG) that increases the diversity of source domain by random augmentation and a Multi-View Domain Classifier (MVDC) that maps features to multiple latent spaces, such that the non-causal factors are removed and the domain-invariant features are purified.
    Extensive experiments on six benchmarks show our MAD obtains state-of-the-art performance.
\end{abstract}

\section{Introduction}
    \label{sec:intro}
    The problem of how to adapt object detectors to unknown target domains in real world has drawn increasing attention.
    Traditional object detection methods \cite{ren2015faster, girshick2015fast, girshick2014rich, redmon2016you, liu2016ssd} are based on independent and identically distributed (i.i.d.) hypothesis, which assume that the training and testing datasets have the same distribution. 
    However, the target distribution can hardly be estimated in real world and differs from the source domains, which is coined as domain shift \cite{5995347}. 
    And the performance of object detection models will sharply drop when facing the domain shift problem. 
    
    Domain adaptation (DA) \cite{chen2018domain, hsu2020every, saito2019strong, xu2020cross,wang2019towards, zhu2019adapting, cai2019exploring} is proposed to deal with the domain shift problem, which enables the model to be adapted to the target distribution by aligning features extracted from the source and unlabeled target domains.
    However, the requirement of target domain datasets still limits the applicability of DA methods in reality.
    Domain generalization (DG) \cite{zhou2022domain} goes one step further, aiming to train a model from single or multiple source domains that can generalize to unknown target domains.

    \begin{figure}[t]
    \centering 
        \includegraphics[width=1\linewidth]{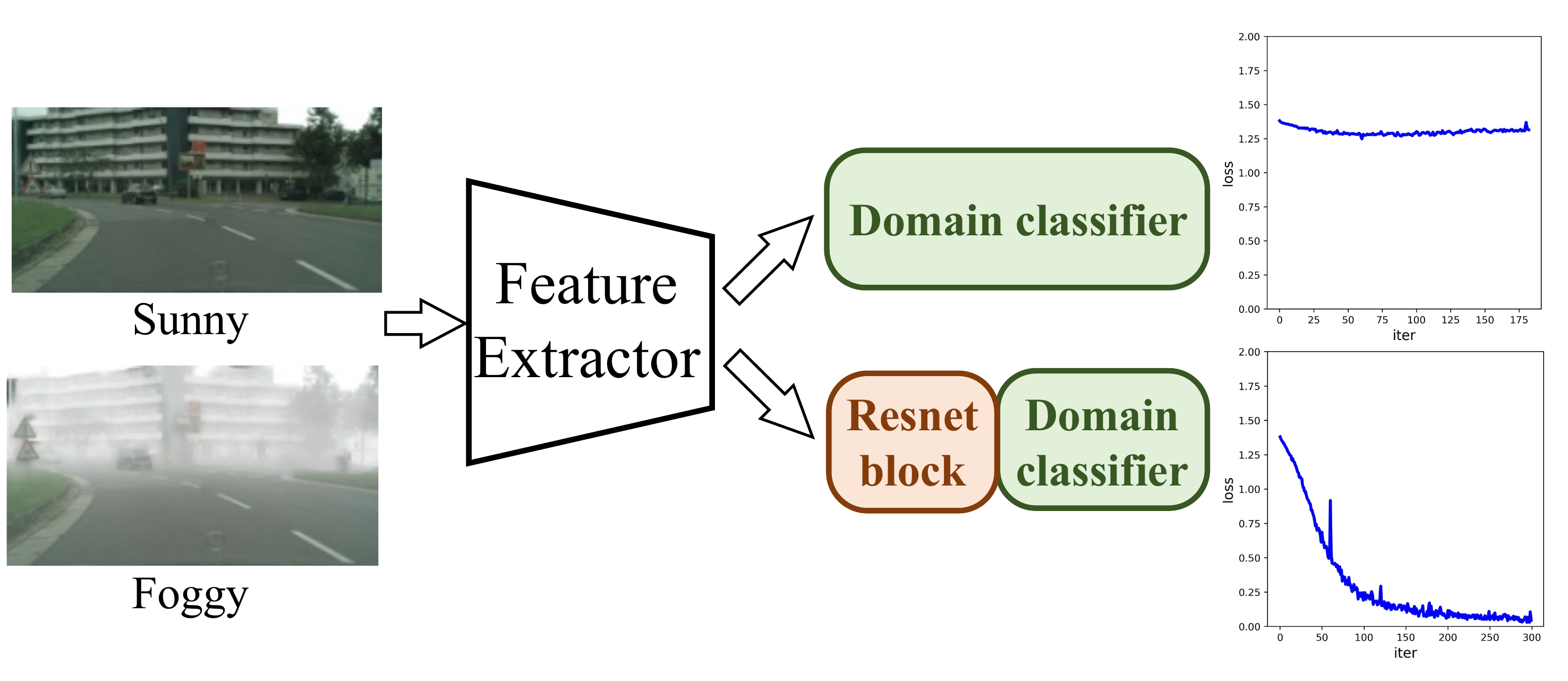}
        \caption{Illustration of the biased learning of conventional DAL. The domain classifier easily encounters early stop and fails.}
        \label{fig:leadin}
    \end{figure}

    Although lots of DG methods have been proposed in the image classification field, there are still some unresolved problems.
    In our opinion, the common features extracted by previous DG methods are still not pure enough.
    The main reason is that through a single-view domain discriminator in DAL, only the significant domain style information can be removed, while some implicit and insignificant non-causal factors in source domains may be absorbed by the feature extractor as a part of common features. This has never been noticed. This implies the multi-mode structure of data and single-view domain discriminator cannot fully interpret the data.
    There is a piece of evidence to support our claim.
  
    To confirm our suspicions on the domain discriminator, we designed a validation experiment.
    As is shown in \cref{fig:leadin}, we use DANN model \cite{ganin2015unsupervised} with DAL strategy to train a common feature extractor.
    When domain classifier converges, we freeze feature extractor and re-train domain classifier with a newly added residual block \cite{he2016deep}. 
    We observe an interesting phenomenon: \textit{when re-trained with the newly added residual block, the domain classifier loss continues to decline}. That is, some domain-specific information still exists. 
    This phenomenon confirms our claim that in existing DG, DAL cannot explore and remove all domain specific features. This is because domain classifier only observes significant domain-specific feature in a single-view, while insignificant domain specific features in one view (space) can be significant in other views (latent spaces).

\begin{figure}[t]
        \centering
        \includegraphics[width=1\linewidth]{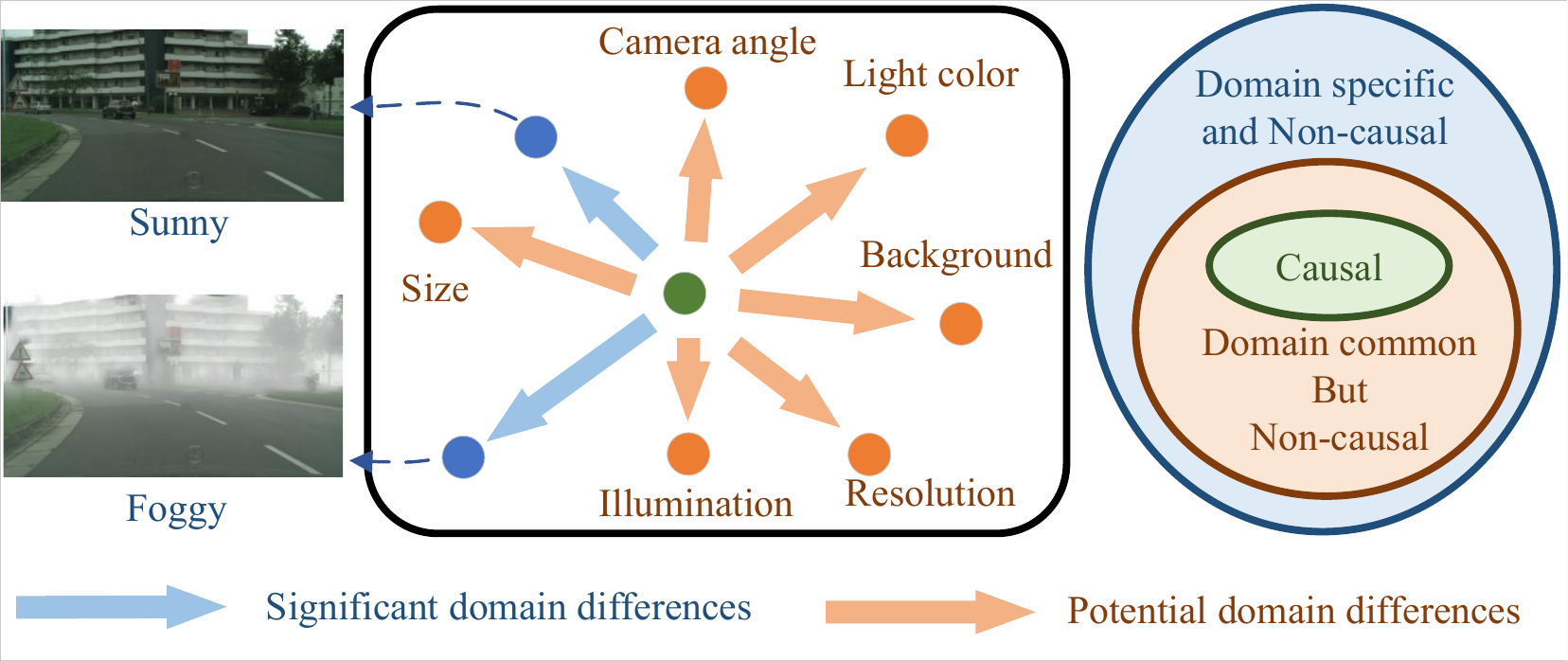}
        \caption{ Relationships among causal factors, noncausal factors, domain specific feature and domain common feature.}
        \label{fig:com-cau}
    \end{figure}

    Based on the former experiment, we propose that  mining common features through DAL in single-view on a limited number of domains is insufficient. By using traditional DAL, only the primary style information w.r.t. domain labels can be removed.
    Here we analyse this problem from the perspective of causality.
    As shown in \cref{fig:com-cau}, in a limited number of domains, the common features still contain non-causal factors such as light color, illumination, background, etc., which is expressed as the orange arrows in the figure.
    And such insignificant non-causal factors observed from one view may still be significant uninformative features in other latent spaces (views). 
    So a natural idea is to explore and remove the implicit non-causal information from multiple views and purify the common features for generalizing to unseen domains.

\begin{figure}[t]
        \centering
        \includegraphics[width=0.9\linewidth]{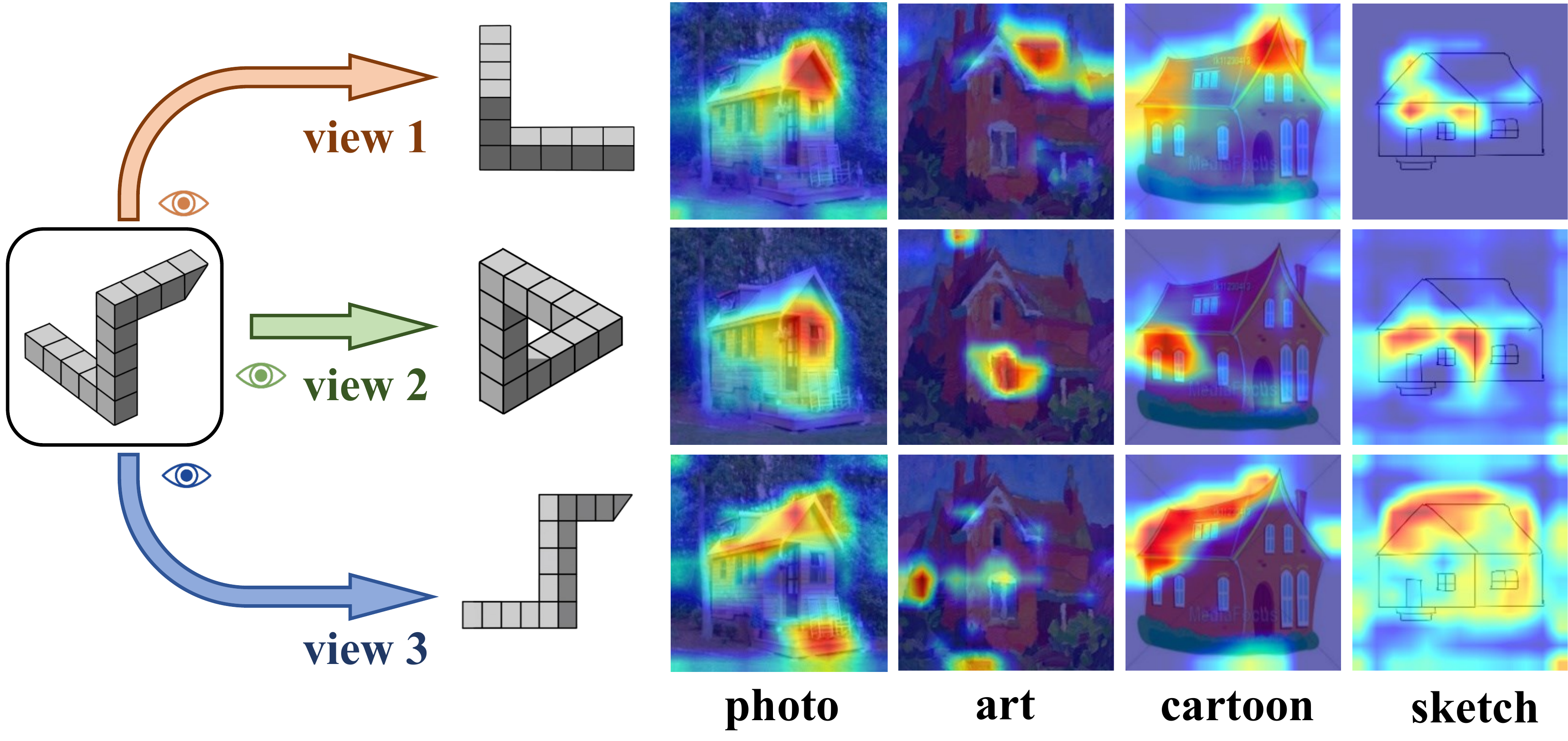}
        \caption{An illustration of the multi-view idea and effect of MAD. Left: a toy example. Right: attention heatmaps of different views.}
        \label{fig:penrose}
    \end{figure}

    In order to remove the potential non-causal information, we rethink the domain discriminator in DAL and propose a multi-view adversarial domain discriminator (MAD) that can observe the implicit insignificant non-causal factors.
    In our life, in order to get the whole architecture of an object, we often need to observe it from multiple views/profiles.
    A toy example is shown in \cref{fig:penrose} (left part). When we observe the Penrose triangle from one specific view, we might misclassify it as a \textit{triangle}, ignoring that it might also appear to be \textit{L} from another perspective.
    Following this intuition, we construct a Multi-View Domain Classifier (MVDC) that can discriminate features in multiple views.
    Specifically, we simulate multi-view observations by mapping the features to different latent spaces with auto-encoders \cite{hinton2006reducing}, and discriminate these transformed features via multi-view domain classifiers.
    By mining and removing as many non-causal factors as possible, MVDC encourages the feature extractor to learn more domain-invariant but causal factors. We conduct an experiment based on MVDC and show the heatmaps from different views in \cref{fig:penrose} (right part), which verifies our idea that different noncausal factors can be unveiled in different views. 

    Although the Multi-View Domain Classifier can remove the implicit non-causal features in principle, it still implies a sufficient diversity of source domains during training. 
    So we further design a Spurious Correlation Generator (SCG) to increase the diversity of source domains.
    Our SCG generates non-causal spurious connections by randomly transforming the low-frequency and extremely high-frequency components, as \cite{huang2021fsdr} points out that in the spectrum of images, the extremely high and low frequency parts contain the majority of domain-specific components. 
    
    Combining MVDC and SCG, the Multi-view Adversarial Discriminator (MAD) is formalized.
    Cross-domain experiments on six standard datasets show our MAD achieves the SOTA performance compared to other mainstream DGOD methods.
    The contributions  are three-fold:
    
    1. We point out that existing DGOD work focuses on extracting common features but fails to mine and remove the potential spurious correlations from a causal perspective.
    
    2. We propose a Multi-view Adversarial Discriminator (MAD) to eliminate implicit non-causal factors by discriminating non-causal factors from multiple views and extracting domain-invariant but causal features.
    
    3. We test and analyze our method on standard datasets, verifying the effectiveness and superiority of our method.

\section{Related Work}
    \label{sec:related-work}
    \subsection{Domain Adaptive Object Detection}
        Object detection is a critical problem in computer vision, aiming to locate and classify the specified instances in specific images.
        Modern object detection methods can be divided into two categories: one-stage methods \cite{redmon2016you, liu2016ssd} and two-stage methods \cite{ren2015faster, girshick2015fast, girshick2014rich}. 
        However, traditional object detection methods suffer from domain shifts in practical applications.
        In order to alleviate the performance degradation caused by domain shift, lots of domain adaptive object detection (DAOD) methods are presented \cite{chen2018domain, hsu2020every, saito2019strong, xu2020cross,wang2019towards, zhu2019adapting, cai2019exploring}.
        DAOD methods are trained with labeled source domains and unlabeled target domains, and alleviate the domain shift problem by DAL.
        The DAOD methods can be divided into two parts: adversarial-based methods and reconstruction-based methods. 
        For the former, the domain adversarial learning structure is introduced to align feature maps by \cite{chen2018domain}. 
        For the latter, \cite{arruda2019cross} firstly uses CycleGAN \cite{zhu2017unpaired} to generate pseudo samples that are similar to the target domain from the source domain samples.
        DAOD methods still have problems in real-world applications.
        On the one hand, they still require additional effort to collect unlabeled target domain datasets, which is expensive and even impossible.
        On the other hand, they cannot guarantee the causality of features. 
        We hope to find domain-invariant but causal features that are more robust for unseen target domains.

    \subsection{Domain Generalization}
        Domain generalization has been studied for a long time in the image classification field.  
        Existing domain generalization methods can be divided into the following three categories.
        First, domain augmentation methods aim to increase the diversity of source domains by transferring images to new domains.
        \cite{shankar2018generalizing, volpi2018generalizing, zhou2020deep} augment source domains in image-space. 
        \cite{xu2021fourier} and \cite{huang2021fsdr} perform augmentation on the frequency spectrum.
        Second, representation learning methods aim to extract domain-invariant representation from source domains. \cite{li2018domain} firstly adopts the idea of DAL in domain adaptation for domain generalization.
        Third, there are also learning strategies like \cite{li2018learning} which firstly adopts meta-learning for domain generalization, following the idea of enabling the network learn how to learn domain-invariant components from different domains.
  
    \subsection{Causal Mechanism}
        Methods based on causal mechanisms \cite{pearl2000models, spirtes2000causation} consider that the prediction based on statistical dependence is unreliable, because the statistical correlations contain both spurious non-causal correlations and causal correlations.
        For example, smoking, yellow teeth and lung cancer are closely related. Nevertheless, only smoking is the causal factor of lung cancer.
        To improve the generalization of methods, they try to mine these invariant causal correlations.
        In recent years, solving DG problems by finding causal factors is gaining more and more attention\cite{peters2016causal, peters2017elements}. 
        Some methods attempt to obtain invariant causal mechanisms \cite{heinze2017conditional, subbaswamy2019preventing, wang2021contrastive}. 
        Meanwhile, other methods try to recover causality characteristics \cite{chang2020invariant, glymour2016causal, liu2021heterogeneous, rojas2018invariant}.
        Existing methods focus on looking for invariant causal factors. 
        However, we argue that one should pay more attention to exploring the potential non-causal spurious correlations, because the domain-invariant representations learned by traditional DAL are often biased towards one view, as \cref{fig:penrose} shows. We propose to purify the domain-invariant features by removing implicit non-causal factors from multiple views in DAL. 

    \begin{figure}
        \centering
            \includegraphics[width=0.85\linewidth]{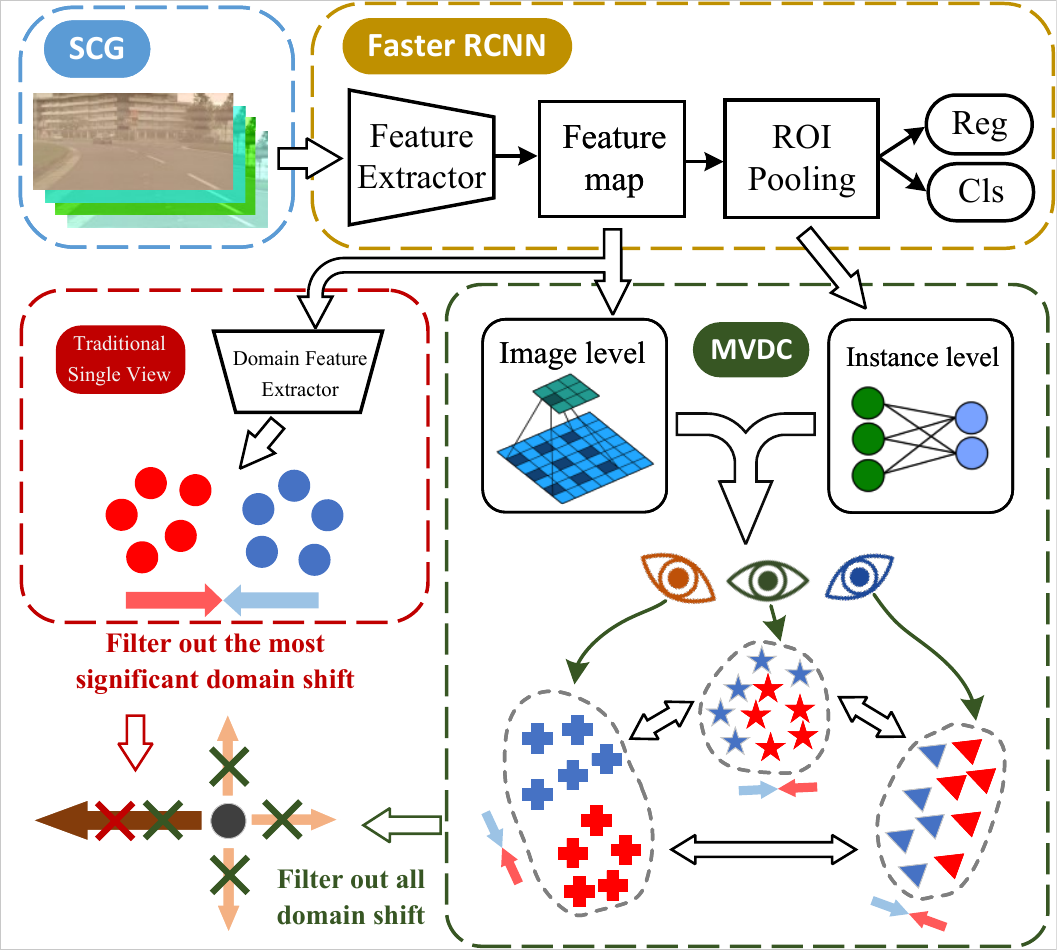}
            \caption{The overall structure of MAD can be divided into three parts.
            (1) {\color{yellow}{Yellow}} part: \textbf{FasterRCNN} backbone network.
            (2) {\color{blue}{Blue}} part: \textbf{SCG} generates potential spurious correlations in the frequency domain.
            (3) {\color{green}{Green}} part: \textbf{MVDC} maps features to different spaces for multi-view DAL and removes implicit non-causal features, such that the domain-invariant but causal features are obtained.
            Notably, the {\color{red}{red}} part shows the conventional DAL.}
        \label{fig:structure}
    \end{figure}
  
    \begin{figure*}
        \centering
            \includegraphics[width=0.9\linewidth]{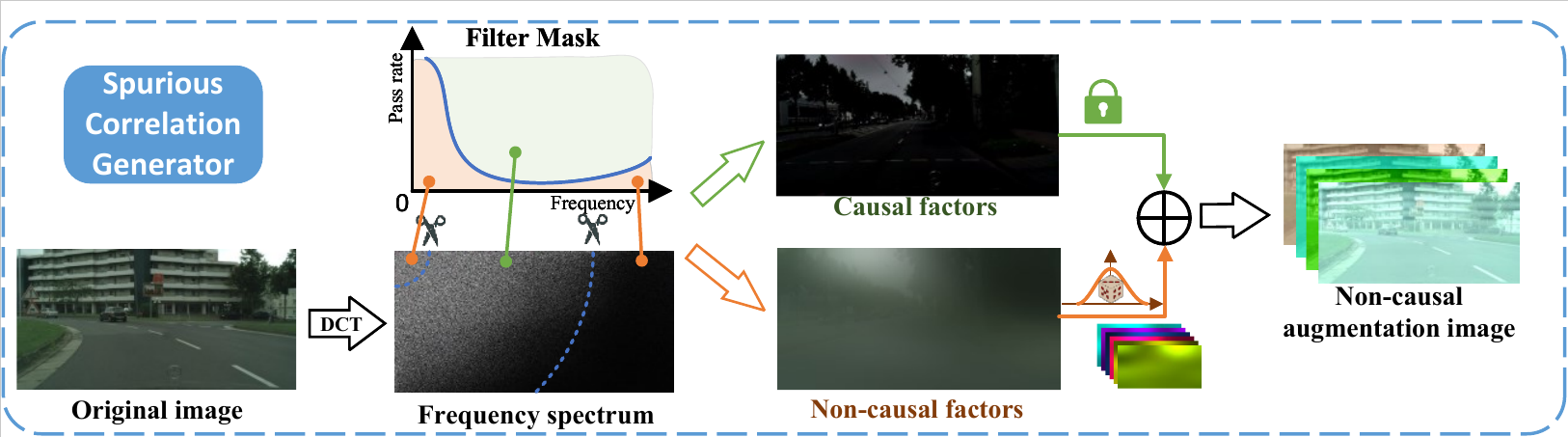}
            \caption{Structure of the Spurious Correlation Generator (SCG).
            The DCT spectrum of the original image is divided into causal and non-causal parts by the band-pass filter.
            The extremely high and low frequency components contain more non-causal factors and the remaining components is considered to contain more causal factors.
            The diversity of source domains is increased by randomising the non-causal part according to a Gaussian distribution.
            After IDCT, an image with potential non-causal factors is generated.}
        \label{fig:SCG}
    \end{figure*}

\section{Proposed MAD Approach}
    \label{sec:methods}
    \subsection{Overview}
        Existing DG methods learn common features with conventional DAL among finite domains \cite{li2018domain, li2018learning}. 
        However, such common features extracted are often not pure due to the implicit non-causal factors.
        As discussed before, we propose a Multi-view Adversarial Discriminator (MAD) to explore and remove potential spurious correlations and encourage the model to extract purer domain-invariant but causal features.
        As is shown in \cref{fig:structure}, our MAD contains two new parts.
        First, a Spurious Correlation Generator (SCG) module is designed to increase the diversity of source domains and make the potential non-causal factors more significant. 
        Second, a Multi-View Domain Classifiers (MVDC) module is designed to identify the non-causal factors for both image and instance levels, such that the domain adversarial learning is more sufficient and the non-causal factors are richer in different views, which instructs the feature extractor to ignore them.
        To summarize, SCG explores and exposes the potential non-causal factors, while MVDC discriminates and removes them.
        
        We make the following definitions to formalize the domain generalization problem.
        The source domain is denoted as $ D_s = \{X_s, Y_s\}$. 
        The feature extractor $f(\cdot)$ can extract the features $S = f(X_s)$ from input images $X_s$. 
        Feature $ S $ contains causal and non-causal factors $\{s_{cau}, s_{non}\}$ in the finite source domains. Intuitively, not all common components $s_{com}$ are causal factors, but domain private components $ s_{pri} $ are non-causal factors, and there is,
        \begin{equation}
            \begin{split}
                \label{eq:re-com-cau}
                \begin{cases}
                s_{cau} \subset s_{com} \\
                s_{non} \supset s_{pri}
                \end{cases}
            \end{split}
        \end{equation}
        The non-causal factors $ s_{non} $ are supposed to obey Gaussian distribution, i.e., $ s_{non}\sim\mathcal{N}(\mu, \sigma^2 ) $ \cite{wang2021regularizing}. 
    \subsection{Spurious Correlations Generator}
        As \cite{huang2021fsdr} pointed out, the extremely high and low frequency parts of images contain more domain-private features. Our SCG aims to increase the diversity of source domains by keeping the causal features invariant and randomizing the non-causal frequency components according to a Gaussian distribution.
        Specifically, our SCG is implemented by the following steps as shown in \cref{fig:SCG}.
        Firstly, by adopting Discrete Cosine Transform ($ \mathscr{F}(\cdot) $) \cite{ahmed1974discrete}, we get the frequency spectrum of the input image $ x \in \mathbb{R}^{H \times W} $ as $ \mathscr{F}(x) $, of which the extremely high and low frequency parts contain more non-causal factors.
        Then, we separate the non-causal factors and causal factors in frequency domain by a band-pass filter: 
        \begin{equation}
            \label{eq:m}
            \begin{split}
                \mathcal{M}(r) = e^{-\frac{u^2+v^2}{{2R_H}^2}} - e^{-\frac{u^2+v^2}{2{R_L}^2}}
            \end{split}
        \end{equation}
        where $ u, v $ denotes the position of the spectrum and $ r(R_L, R_H) $ denote the cut-off frequency of low and high frequency.
        We then randomize this non-causal factor $S$ according to a Gaussian distribution as $ R_{G}(S) = S \cdot (1 + \mathcal{N}(0,1)) $. 
        Finally, we get the augmented image $\hat{x}$ with potential non-causal factors by adopting the Inverse Discrete Cosine Transform $ \mathscr{F}^{'}(\cdot ) $ to the augmented spectrum. 
        Our spurious correlations generator can be expressed as: 
        \begin{equation}
            \begin{split}
                \label{eq:aug} 
                \hat{x} = \mathscr{F}^{'}( R_{G}(\mathcal{M}(r) \cdot \mathscr{F}(x)) + (1-\mathcal{M}(r)) \cdot \mathscr{F}(x) )
            \end{split}
        \end{equation}

    \subsection{Multi-View Domain Classifier}
        \subsubsection{Domain Adversarial Learning (DAL)}
        
        DAL \cite{ganin2015unsupervised} is a standard method to extract the common feature of different domains, which minimizes the $ \mathcal{A} -Distance $ of the extracted features between different domains. 
        DAL is a minimax optimization problem between feature extractor $\mathcal{F}$ and the ideal domain classifier: 
        \begin{equation}
            \begin{split}
                \label{eq:minmax}
                \min_{\mathcal{F}}d_{\mathcal{A}}(D_{s1},D_{s2})=&\underbrace{\max_{\mathcal{F}}\min_{h \in \mathcal{H}}err(h(s))}_{Standard \ DAL} \\
                \Rightarrow&\underbrace{\max_{\mathcal{F}}\sum_{i=1}^{M}\min_{h_i \in \mathcal{H}, e_i}err(h_i(e_i(s)))}_{Ours}
            \end{split}
        \end{equation}
        where $ \mathcal{H} $ denotes a hypothesis set of all possible domain classifiers, $ h(\cdot) $ is one of the domain classifiers in $ \mathcal{H} $, and $e(\cdot)$ denotes encoders which map feature to divers latent spaces.
        A single $ h $ depends on the most discriminative domain private features, so it ignores the insignificant domain specific components of the features and incorrectly takes such non-causal components as common features.
    
        We, therefore, propose to improve the sensitivity of the domain classifier to potential non-causal factors by extending DAL to more views. 
        Specifically, our MVDC can map features into multiple latent spaces with encoders $ e_i $ and then discriminate features in each space with an independent domain classifier $ h_i $. 
        These domain classifiers encourage the feature extractor $\mathcal{F}$ to ignore the implicit non-causal factors and learn domain-invariant but causal features. 

        \begin{figure}[t]
            \centering
                \includegraphics[width=0.9\linewidth]{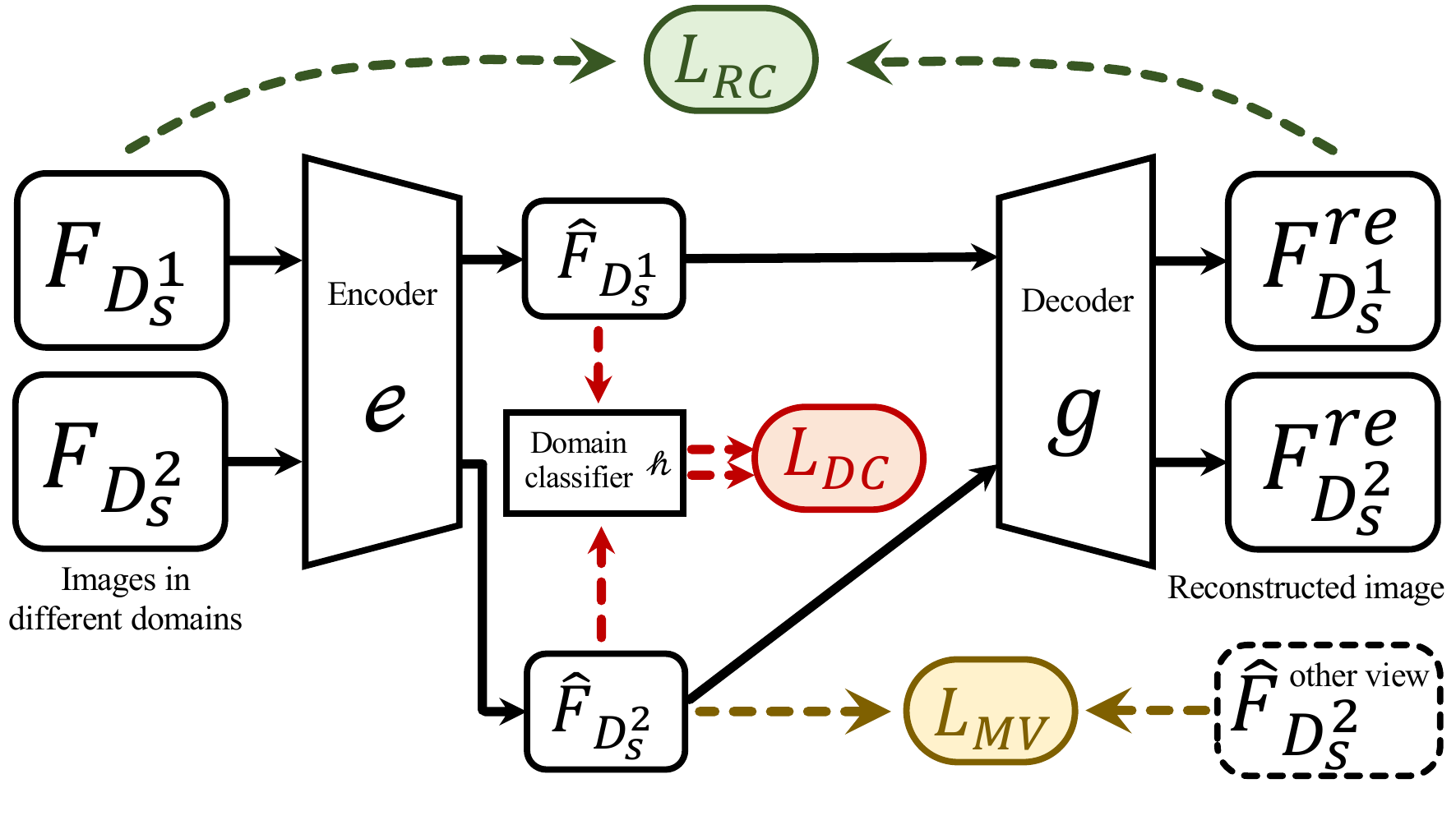}
                \caption{Structure of one branch of MVDC.
                It contains three constraints:
                $\mathcal{L}_{RC}$ ensures the mapped features contain complete semantic information.
                $\mathcal{L}_{DC}$ makes the domain discriminator of each view have domain classification ability.
                $\mathcal{L}_{MV}$ ensures that each auto-encoder maps features to different latent spaces.}
                \label{fig:5}
        \end{figure}
        \subsubsection{Classifier Structure}
        \cref{fig:5} shows the structure of one branch of the MVDC, which represents one of the multiple views to observe the features.
        The complete structure of MVDC contains $M$ branches for image-level features and $M$ branches for instance-level features respectively. 
        Each branch of MVDC contains an auto-encoder and a classifier in structure.
    
        The encoder and decoder are the basic network structure of each branch, which map features into different latent spaces to show different profiles of the feature.
        The encoder part aims to compress the features and map them into different latent spaces. 
        Then the latent features are fed into an independent domain classifier.
        Meanwhile, in order to ensure the semantic content invariance of features, the latent features are mapped back into the original space through a subsequent decoder.
        
        To explore the non-causal factors hidden in the whole image and each instance, we make different designs on multi-view domain classifiers for image-level and instance-level respectively.
        For the image-level, we focus on the global non-causal factors of an image, such as the illumination, color and background texture. 
        These global non-causal factors are similar across the image, so we use the convolutional layers to construct the encoder and decoder. 
        In each branch, we use dilated convolution \cite{yu2015multi} with different dilation rates to extract different non-causal factors of domains.
        For the instance-level, we use fully connected layers to mind more semantic non-causal factors like the camera angle of each instance.

    \begin{table*}
      \centering
      \setlength{\tabcolsep}{4mm}{
      \begin{tabular}{ c|c|c c c c }
        \hline
        \diagbox{Source}{Target}          & Method                & Cityscapes    & Foggy Cityscapes  & Rain Cityscapes   & BDD100k       \\ \hline \hline
        \multirow{6}{*}{Cityscapes}       & Source-only           & ---            & 27.2              & 36.3              & 24.0          \\ 
                                          & MLDG                  & ---            & 29.2              & 42.1              & 21.0          \\ 
                                          & FACT                  & ---            & 25.3              & 39.9              & 26.0          \\ 
                                          & FSDR                  & ---            & 31.0              & \textbf{42.8}     & 26.2          \\ 
                                          & DANN+\textbf{SCG}     & ---            & 37.5              & 39.1              & 26.1          \\ 
                                          & \textbf{MAD(Ours)}   & ---            & \textbf{38.6}     & 42.3              & \textbf{28.0} \\
        \hline
        \multirow{6}{*}{Foggy Cityscapes} & Source-only           & 29.9          & ---               & 38.4              & 17.5          \\ 
                                          & MLDG                  & 30.4          & ---               & 38.6              & 18.0          \\ 
                                          & FACT                  & 30.0          & ---               & 38.7              & 20.2          \\ 
                                          & FSDR                  & 31.3          & ---               & 40.8              & 20.4          \\ 
                                          & DANN+\textbf{SCG}     & 38.4          & ---               & 40.4              & 22.4          \\ 
                                          & \textbf{MAD(Ours)}   & \textbf{41.3} & ---               & \textbf{43.3}     & \textbf{24.4} \\
        \hline
        \multirow{6}{*}{BDD100k}          & Source-only           & 33.6          & 27.2              & 34.3              & ---           \\ 
                                          & MLDG                  & 24.7          & 17.1              & 20.0              & ---           \\ 
                                          & FACT                  & 32.4          & 24.3              & 33.9              & ---           \\ 
                                          & FSDR                  & 32.4          & 27.8              & 34.7              & ---           \\ 
                                          & DANN+\textbf{SCG}     & 35.8          & 29.3              & 33.9              & ---           \\ 
                                          & \textbf{MAD(Ours)}   & \textbf{36.4} & \textbf{30.3}     & \textbf{36.1}     & ---           \\
        \hline
      \end{tabular}}
      \caption{Results on four domains (\textbf{C}, \textbf{F}, \textbf{R}, \textbf{B}) trained on single source domain.
      Note that during training, the target domain is unseen according to DG setting. The best mAP are highlighted in bold. 
      }
      \label{tab:ce}
    \end{table*}

        \subsubsection{Loss Function} 
        For training the object detector with MAD approach, several loss functions are introduced in the following text.
        
        First, a reconstruction loss is used to ensure that the semantics of features are not changed by the encoder.
        The mapped feature $ e(s) $ should contain the semantic information required to reconstruct the original feature $ s $.
        Only if the semantic information is guaranteed to be complete, the subsequent domain classifier is meaningful.
        So we use MSE loss to constrain the distance between the original feature $ s $ and the reconstructed feature $ g(e(s)) $. The reconstruction loss can be described as:
        \begin{equation}
            \begin{split}
                \mathcal{L}_{RC}=\frac{1}{M}\sum_{m=1}^{M}MSE(s, g_m(e_m(s))) 
                \label{eq:L_RC}
            \end{split}
        \end{equation}
    
        Second, the adversarial domain classifier loss is used to ensure the mapped features are domain distinguishable (inner optimization) and domain confused (outer optimization).
        We use Cross-Entropy loss to adversarially train the K-domain classifiers in total $ M $ branches.
        And the domain label in $k^{th}$ domain is denoted as $ y_k $.
        \begin{equation}
            \begin{split}
                \mathcal{L}_{DC}= -\frac{1}{M} \sum_{m=1}^{M}\sum_{k=1}^{K} y_k \cdot log(p(D_m(e_m(s_k))))
                \label{eq:L_DC}
            \end{split}
        \end{equation}
    
        The third constraint is the most critical view-different loss, which ensures the auto-encoders to map features into diverse latent spaces.
        Therefore, we propose to enlarge the feature difference between latent spaces (views), such that the insignificant non-causal factors become significant.
        So we construct the following MSE loss of each feature pair from $M$ different latent spaces.
        \begin{equation}
            \begin{split}
                \mathcal{L}_{MV}=-\frac{\sum_{i}^{M}\sum_{j,i \neq j}^{M}||e_i(s) - e_j(s)||^2}{M^2-M} 
                \label{eq:L_MV}
            \end{split}
        \end{equation}

        The fourth constraint is used to ensure the consistency of the results in these $2M$ branches of two levels.
        For each pair of image-level and instance-level branches, we adopt $l_2$ distance between the average value of image-level predictions $ p_{i}^{(u,v)} $ and each instance-level prediction $ p_{j,n} $ as the consistency constraint. 
        We suppose the feature map of each image contains $|I|$ pixels and $N$ instances in total.
        Then, the consistency loss of the whole model can be described as:
        \begin{equation}
            \begin{split}
                \mathcal{L}_{cst} = \sum_{i,j}^{M}\sum_{n}^{N}\|\frac{1}{|I|} \sum_{u,v}p_{i}^{(u,v)}-p_{j,n}\|_2
                \label{eq:cst}
            \end{split}
        \end{equation}
    
        The MVDC loss in both image level and instance level can be presented as:
        \begin{equation}
            \begin{split}
                \mathcal{L}_{MVDC}^{(img,ins)} = \mathcal{L}_{RC} + \mathcal{L}_{DC} + \mathcal{L}_{MV}
                \label{eq:MVDC}
            \end{split}
        \end{equation}
        
        Then, we obtain the overall loss of MAD by trade-off the object detector loss and MVDC loss with $\lambda$ as:
        \begin{equation}
            \begin{split}
                \mathcal{L}_{MAD} = \mathcal{L}_{det} + \lambda(\mathcal{L}_{MVDC}^{img} + \mathcal{L}_{MVDC}^{ins} + \mathcal{L}_{cst})
                \label{eq:all}
            \end{split}
        \end{equation}

\section{Experiments}
    \label{sec:Exp}
    \subsection{Datasets}
    We adopt seven cross-domain object detection benchmark datasets, which will be introduced below.
    Cityscapes \cite{cordts2016cityscapes} dataset mainly contains daytime scenery in the streets and Foggy Cityscapes \cite{cordts2016cityscapes} and Rain Cityscapes \cite{Hu_2019_CVPR} are datasets of synthesized images with different weather conditions based on the depth information from the Cityscapes.
    SIM10k~\cite{johnson2016driving} dataset contains rendered images of rendered 3D models.
    KITTI \cite{geiger2013vision} is an autonomous driving dataset. 
    PASCAL VOC \cite{everingham2015pascal} dataset is collected from the real world. 
    The BDD100k~\cite{yu2020bdd100k} dataset is a large-scale dataset for autonomous driving.
    We abbreviate \{SIM 10k, Cityscapes, Foggy Cityscapes, Rain Cityscapes, BDD100k, KITTI, PASCAL VOC\} as \{S, C, F, R, B, K, V\} respectively in the following text.
  
    \subsection{Experimental Setup}
    
        \begin{table*}
          \centering
          \begin{tabular}{p{0.5mm} l|c|c c c c c c c c|c }
              \hline
              \multicolumn{2}{ l|}{Methods}                                       & Dataset used      & person        & rider         & car           & truck         & bus           & train         & motor         & bike          & mAP \\
              \hline
              \multicolumn{2}{ l|}{Source-only}                                   & Single Source     & 27.1          & 39.3          & 36.0          & 14.2          & 31.4          & 9.4           & 26.9          & 33.4          & 27.2 \\
              \hline
              \multicolumn{1}{ l|}{\multirow{8}*{DA}}& DAF\cite{chen2018domain}   &                   & 31.6          & 43.6          & 42.8          & 23.6          & 41.3          & 21.2          & 28.9          & 32.6          & 33.2 \\
              \multicolumn{1}{ l|}{}& SW-DA\cite{saito2019strong}                 &                   & 31.8          & 44.3          & 48.9          & 21.0          & 43.8          & 28.0          & 28.9          & 35.8          & 35.3 \\
              \multicolumn{1}{ l|}{}& SC-DA\cite{zhu2019adapting}                 & Single Source     & 33.8          & 42.1          & 52.1          & 26.8          & 42.5          & 26.5          & 29.2          & 34.5          & 35.9 \\
              \multicolumn{1}{ l|}{}& MTOR\cite{cai2019exploring}                 & \&                & 30.6          & 41.4          & 44.0          & 21.9          & 38.6          & 40.6          & 28.3          & 35.6          & 35.1 \\
              \multicolumn{1}{ c|}{}& ICR-CCR\cite{xu2020exploring}               & Target images     & 32.9          & 43.8          & 49.2          & 27.2          & 45.1          & 36.4          & 30.3          & 34.6          & 37.4 \\
              \multicolumn{1}{ l|}{}& Coarse-to-Fine\cite{zheng2020cross}         & (without labels)  & 34.0          & \textbf{46.9} & 52.1          & \textbf{30.8} & 43.2          & 29.9          & 34.7          & 37.4          & 38.6 \\
              \multicolumn{1}{ l|}{}& GPA\cite{xu2020cross}                       &                   & 32.9          & 46.7          & 54.1          & 24.7          & \textbf{45.7} & \textbf{41.1} & \textbf{32.4} & \textbf{38.7} & 39.5 \\
              \multicolumn{1}{ l|}{}& Center-Aware\cite{hsu2020every}             &                   & \textbf{41.5} & 43.6          & \textbf{57.1} & 29.4          & 44.9          & 39.7          & 29.0          & 36.1          & \textbf{40.2} \\
              \hline
              \hline
              \multicolumn{1}{ l|}{\multirow{3}*{DG}}& DIDN\cite{lin2021domain}   & Multiple Source   & 31.8          & 38.4          & \textbf{49.3} & \textbf{27.7} & 35.7          & 26.5          & 24.8          & 33.1          & 33.4 \\
              \cline{2-12}
              \multicolumn{1}{ l|}{}& LMDG\cite{li2018learning}    & \multirow{4}*{Single Source}     & 32.2          & 41.7          & 38.9          & 19.2          & 33.0          & 9.1           & 23.5          & 36.3          & 29.2 \\
              \multicolumn{1}{ l|}{}& FACT\cite{xu2021fourier}                    &                   & 26.2          & 41.2          & 35.9          & 13.6          & 27.7          & 3.0           & 23.3          & 31.3          & 25.3 \\
              \multicolumn{1}{ l|}{}& FSDR\cite{huang2021fsdr}                    &                   & 31.2          & 44.4          & 43.3          & 19.3          & 36.6          & 11.9          & 27.1          & 34.1          & 31.0 \\
              \multicolumn{1}{ l|}{}& \textbf{MAD}                               &                   & \textbf{34.2} & \textbf{47.4} & 45.0          & 25.6          & \textbf{44.0} & \textbf{42.4} & \textbf{30.28}& \textbf{40.12}& \textbf{38.6} \\
              \hline
              \multicolumn{2}{ l|}{Oracle - Train on target}                      & Target            & 37.8          & 47.4          & 53.0          & 31.6          & 52.9          & 34.3          & 37.0          & 40.6          & 41.8 \\
              \hline
          \end{tabular}
          \caption{Results of DG and DA experiments tested on Foggy Cityscapes (\textbf{F}).
          We compare our MAD method with typical DAOD methods and DGOD methods.
          DAOD methods are trained on \textbf{C} and unlabeled \textbf{F}, 
          multi-domain DGOD methods are trained on \textbf{C} and \textbf{B}, 
          and single-domain DGOD methods are trained on \textbf{C}.
          The best AP in each class and mAP are highlighted in bold.
          }
          \label{tab:mm}
        \end{table*}
        
        \begin{table}
            \centering
            \begin{tabular}{l|c c c c c c }
                \hline
                Method               & \textbf{F}    & \textbf{R}   & \textbf{B}    & \textbf{V}    & \textbf{S}    & \textbf{K}    \\
                \hline
                SourceOnly           & 36.0          & 39.0         & 41.3          & 62.0          & 39.2          & 73.4          \\
                DAF                  & 42.8          & 52.9         & 41.4          & 59.2          & 39.0          & 72.1          \\
                MLDG                 & 38.9          & 52.7         & 39.4          & 61.4          & 37.2          & 63.9          \\
                FACT                 & 35.9          & 48.8         & 42.0          & 65.3          & 41.2          & 73.2          \\
                FSDR                 & 43.3          & 52.7         & \textbf{45.4} & 63.4          & 42.2          & 73.8          \\
                \textbf{MAD}        & \textbf{45.0} & \textbf{54.0}& 42.4          & \textbf{67.6} & \textbf{43.2} & \textbf{74.1} \\
                \hline
            \end{tabular}
            \caption{Results of SourceOnly, DAF, MLDG, FACT, FSDR and our MAD from \textbf{C} to \textbf{F}, \textbf{B}, \textbf{V}, \textbf{S} and \textbf{K} on shared category \{\emph{car}\}.
            We train DGOD methods with single source domain \textbf{C}, while using both \textbf{C} and unlabeled \textbf{F} in the training process of DAF.
            The best mAP is highlighted in bold.
            }
            \label{tab:md}
        \end{table}

        \textbf{Implementation Details.}
        First, to verify the effectiveness of our MAD method, we conduct cross-test experiments on \{\textbf{C}, \textbf{F}, \textbf{R}, \textbf{B}\}, which means we train a model on one of these datasets and test it on the rest datasets.
        For each source and target pair, we only calculate the result on the intersection of their label space.
        To uniform the annotation styles of these datasets, we regard labels \{\emph{motor}, \emph{motorcycle} and \emph{motorbike}\} as \{\emph{motor}\} and labels \{\emph{bike} and \emph{bicycle}\} as \{\emph{bike}\}.
        
        Second, to verify the superiority of our MAD method, we compare with the existing DGOD methods like MLDG \cite{li2018learning}, DIDN \cite{lin2021domain}, FACT~\cite{xu2021fourier} and FSDR~\cite{huang2021fsdr}.
        Several DAOD methods such as DAF \cite{chen2018domain}, SW-DA \cite{saito2019strong}, SC-DA \cite{zhu2019adapting}, MTOR \cite{cai2019exploring}, GPA \cite{xu2020cross} are also compared under the task from cityscapes to foggy cityscapes. 
        We train a total of 10 epochs.
        In training process, we set the initial learning rate to 0.002, and start to attenuate the learning rate to 0.0002 at the 7th epoch to make the model converge better.
        In our experiments, we train the models with MindSpore\cite{chen2021deep} and PyTorch frameworks. 
        Our code is available at \href{http://github.com/K2OKOH/MAD}{github.com/K2OKOH/MAD}.
        Mean average precisions (mAP) with a IoU threshold of 0.5 is reported.
    
        \textbf{Baseline.}
        We build our method on the basis of FasterRCNN \cite{ren2015faster} framework with vgg16 pre-trained on ImageNet\cite{russakovsky2015imagenet} as the backbone and adopt the Stochastic Gradient Descent (SGD)\cite{robbins1951stochastic} as the optimization method.

        \begin{figure}[t]
            \centering
            \includegraphics[scale=0.3]{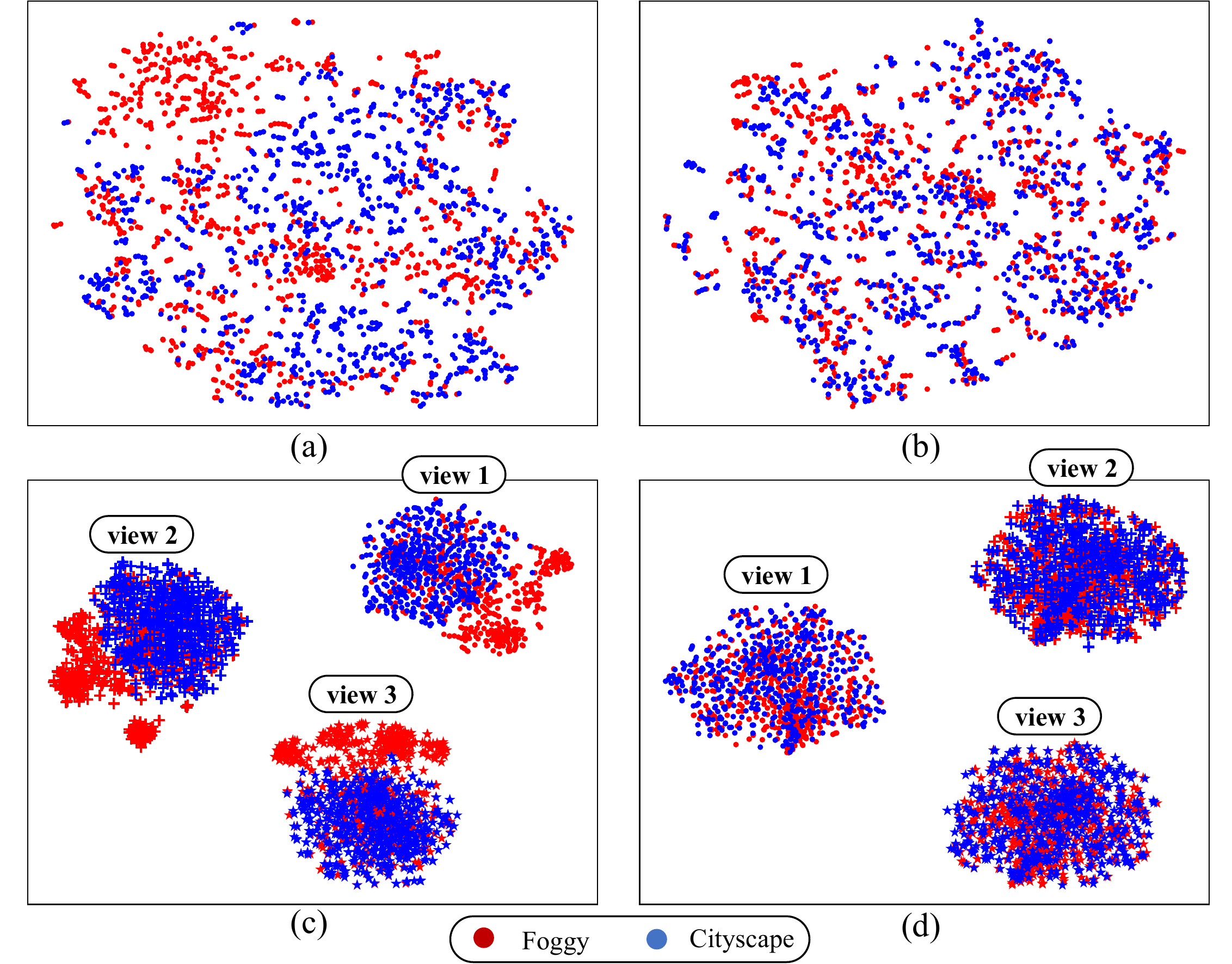}
            \caption{Visualization of the feature distribution via t-SNE. 
            (a) Result of Faster RCNN.
            (b) Result of DANN under single-view.
            (c) Result of DANN under multi-view.
            (d) Result of MAD.
            }
            \label{fig:TSNE}
        \end{figure}

    \subsection{Results and Discussion}

        The results in \cref{tab:ce} show that our method can achieve better results in most cross-domain scenarios.
        Trained with limited number of source domains, our SCG method can add non-causal factors in more directions to the existing images, which better simulate the potential target domain distribution. 
        This makes our method superior to MLDG, FACT and FSDR that extract features over finite known source domains or fixed augmented domains.
        Comparing the single-view DANN with our MAD, we can also find that our MAD performs better, which shows that our Multi-View domain classifier can further help mine and remove non-causal factors from the simulated target distribution.

        In \cref{tab:mm}, we compare our MAD with mainstream DG and DA methods.
        Under both single-source and multi-source DG settings, our method has the best generalization ability among DG methods and exceeds Multi-Source methods in most categories.
        Furthermore, our target-free MAD can even surpass some of the domain adaptive methods, which are trained with unlabeled target images.

        We further conduct experiments on the common categories \textit{car} in six datasets (\textbf{C}, \textbf{F}, \textbf{R}, \textbf{S}, \textbf{K}, \textbf{V}  and \textbf{B}) to verify the domain generalization ability of our MAD.
        As shown in \cref{tab:md}, our method also performs the best in most unseen target domains.

        As is shown in \cref{fig:TSNE}, we also perform visualization of feature distribution via t-SNE under the task from \textbf{C} to \textbf{F}.
        (a) shows the feature distribution of cars in datasets \textbf{C} and \textbf{F} extracted with the original Faster RCNN model, in which the difference of distribution between domains is clear. 
        (b) shows the feature distribution of the same datasets and category extracted by DANN, from which we can see that DANN can align the distribution of different domains in a single view.
        However, as shown in (c), the aligned feature distributions by DANN are still separated with multi-view discriminators by our MVDC, which means that DANN can only remove the significant non-causal factors and the remained insignificant non-causal factors are still domain discriminative.
        Compared to (c), (d) shows the feature distribution in multi-view extracted by our MAD, and we can see that our MAD can indeed map features into different spaces and well-align different domains under each view.

        The multi-view adversarial discriminator is substantially orthogonal to the computer vision tasks and thus can also be applicable to DG-based image classification tasks. 
        Therefore, we conduct single-source DG experiments on the widely used PACS and VLCS datasets, and compared with ERM and DANN frameworks, as shown in \cref{tab:IC}. The results show the effectiveness of our MAD.

         \begin{table}
            \centering
            \resizebox{\columnwidth}{!}{
            \setlength{\tabcolsep}{1mm}{
                \begin{tabular}{c|c c c c|c c c c|c c c c|c c c c}
                \hline
                Source  & \multicolumn{16}{c}{Target} \\
                \hline
                        & \multicolumn{4}{c}{ERM}          & \multicolumn{4}{|c}{ERM+SCG}     & \multicolumn{4}{|c}{DANN+SCG}      & \multicolumn{4}{|c}{MVDC+SCG (MAD)}              \\    
                \hline
                \hline
                PACS    & P      & A        & C         & S     & P     & A     & C     & S     & P     & A         & C         & S     & P             & A                 & C                 & S                 \\
                \hline
                P       &  -     & 61.9     & 26.2      & 31.9  & -     & 62.8  & 29.3  & 40.1  & -     & 63.1      & 35.3      & 43.1  & -             & \textbf{66.6}     & \textbf{40.9}     & \textbf{44.2}     \\
                A       & 90.6   & -        & 67.3      & 57.2  & 90.8  & -     & 68.7  & 61.7  & 91.4  & -         & 70.7      & 64.3  & \textbf{92.6} & -                 & \textbf{71.2}     & \textbf{68.9}     \\
                C       & 79.5   & 64.1     &  -        & 65.6  & 78.6  & 64.3  & -     & 69.0  & 79.2  & 63.6      & -         & 69.3  & \textbf{79.9} & \textbf{64.6}     & -                 & \textbf{70.9}     \\
                S       & 48.0   & 42.8     & 60.5      & -     & 49.4  & 51.5  & 62.2  & -     & 48.7  & 53.8      & 63.4      & -     & \textbf{53.2} & \textbf{57.4}     & \textbf{63.8}     & -                 \\
                \hline
                \hline
                VLCS    & V      & L        & C         & S     & V     & L     & C     & S     & V     & L         & C         & S     & V             & L                 & C                 & S                 \\
                \hline
                V       &  -     & 39.6     & 96.1      & 68.9  & -     & 40.1  & 97.6  & 69.2  & -     & 43.4      & 98.3      & 69.5  & -             & \textbf{47.2}     & \textbf{98.5}     & \textbf{71.4}     \\
                L       & 61.3   & -        & 82.6      & 43.8  & 61.7  & -     & 83.7  & 46.9  & 61.7  & -         & 83.7      & 46.9  & \textbf{62.2} & -                 & \textbf{86.7}     & \textbf{51.8}     \\
                C       & 50.6   & 20.7     &  -        & 42.7  & 51.2  & 21.9  & -     & 43.5  & 51.7  & 27.2      & -         & 44.9  & \textbf{51.8} & \textbf{29.6}     & -                 & \textbf{46.0}     \\
                S       & 60.2   & 45.5     & 72.7      & -     & 60.9  & 47.4  & 72.9  & -     & 62.4  & 50.0      & 74.9      & -     & \textbf{64.0} & \textbf{51.3}     & \textbf{75.4}     & -                 \\
                \hline
            \end{tabular}
            }}
            \caption{Classification results on PACS and VLCS datasets.}
            \label{tab:IC}
        \end{table}

 \begin{table}
          \centering
          \begin{tabular}{ c c c c|c c c }
            \hline
            \multicolumn{4}{c|}{\textbf{Methods}}                           &               & \textbf{mAP}    &              \\
            SCG           & INS           & IMG             & CST           & C to F        & C to R          & C to B       \\
            \hline
                          &               &                 &               & 27.2          & 36.3            & 24.0         \\
            \checkmark    &               &                 &               & 34.1          & 37.9            & 25.4         \\
            \checkmark    & \checkmark    &                 &               & 38.2          & 40.7            & 26.7         \\
            \checkmark    & \checkmark    & \checkmark      &               & 38.3          & 41.0            & 26.2         \\
            \checkmark    & \checkmark    & \checkmark      & \checkmark    & \textbf{38.6} & \textbf{42.3}   & \textbf{28.0}  \\
            \hline
          \end{tabular}
          \caption{Ablation results of each component in MAD trained on \textbf{C} and tested on \textbf{F}, \textbf{R}, \textbf{B}.
          \underline{SCG}: Spurious Correlations Generator.
          \underline{IMG}: Image-level Multi-View Discriminator.
          \underline{INS}: Instance-level Multi-View Discriminator.
          \underline{CST}: Consistent loss of img and ins.}
          \label{tab:ablation1}
        \end{table}

    \subsection{Ablation Study}

        We conducted an ablation study on our MAD methods to verify the validity of each part.
        Our method can be divided into four parts in total, namely spurious correlations generator (SCG), image-level and instance-level multi-view domain classifier (IMG, INS) and the consistency constraints (CST).
        We study the contribution of each part by adding them sequentially and observing the change in mAP performance.
        We train MAD on domain \textbf{C} and test it on other domains \textbf{F}, \textbf{R}, \textbf{B} to conduct ablation experiments.
    
        \cref{tab:ablation1} reflects the effectiveness of each part of our MAD.
        By introducing SCG, potential spurious correlations are injected into the network.
        The MVDC consisting of three submodules (IMG, INS, CST) further mines and removes insignificant spurious correlations in the domains. 
        Specifically, the image-level adversarial submodule (IMG) eliminates overall non-causal factors, and the instance-level submodule (INS) eliminates the semantic non-causal factors in each instance.
        The consistency loss (CST) ensures the consistency of the domain discriminators in two stages.
  
    \subsection{Hyper-parameters Analysis}
        We tested two hyper-parameters of our MAD method.
        
\begin{figure}[t]
          \centering
            \includegraphics[scale=0.32]{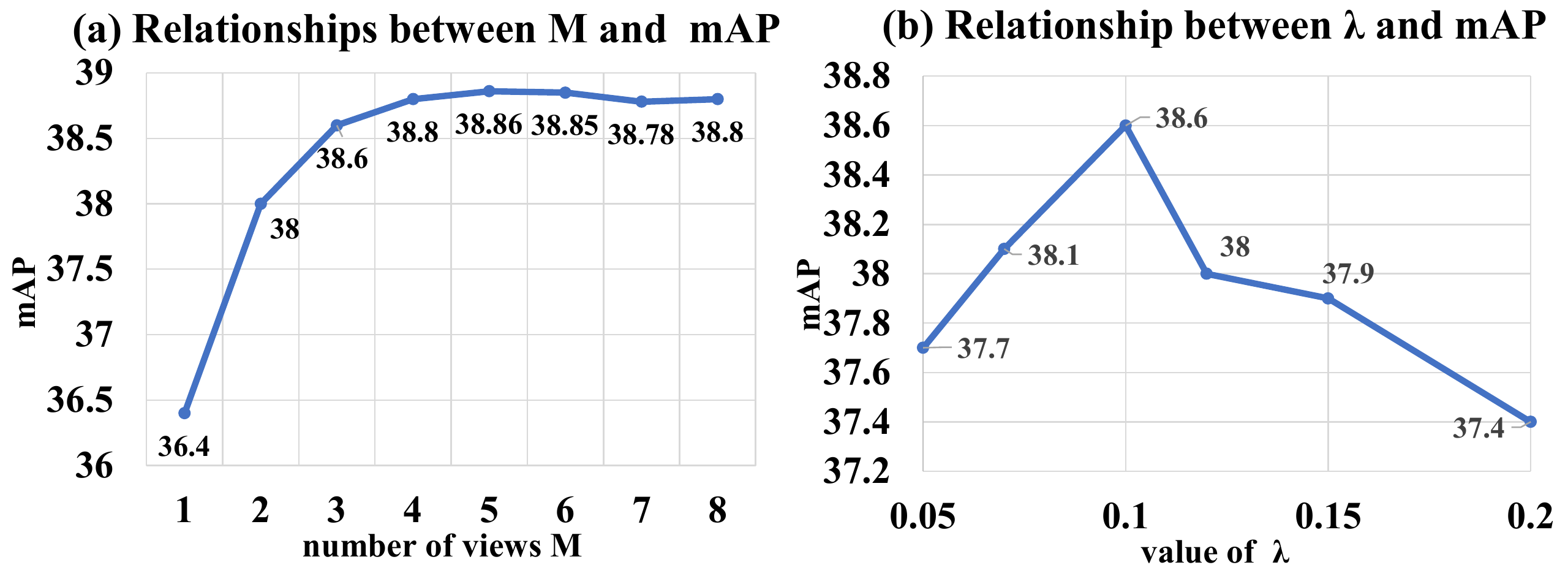}
            \caption{Ablation results of hyperparameters. 
            (a) reflects the performance w.r.t. the number of views (branches).
            (b) reflects the performance w.r.t. the trade-off coefficient $ \lambda $.}
          \label{fig:n-m}
        \end{figure}

        First, the number $ M $ of views is the key hyper-parameter in MAD. 
        More views lead to better performance, but too many auto-encoders increase model complexity with diminishing marginal effect.
        As we can see in \cref{fig:n-m} (a), we found that performance improved until $M=5$ and then converged with further views until $M=8$.
        Thus, we set $M=3$ as a balance between performance and cost.
        
        Second, the trade-off coefficient $ \lambda $ of the domain adversarial loss in \cref{eq:all} is used to balance the main task of object detection and the MVDC part.
        We take several values from 0.05 to 0.2 for testing. 
        As can be seen from \cref{fig:n-m} (b), we set $ \lambda=0.1 $ in MAD for all the experiments.
            
\section{Conclusion}
    \label{sec:Ccs}
    This paper analyzes the problem of domain adversarial learning (DAL) from the perspective of causal mechanisms.
    We point out that existing DG methods fail to remove potential non-causal factors implied in common features, because DAL is biased by the single-view nature of the domain discriminator.
    To overcome this problem, we propose a Multi-view Adversarial Discriminator (MAD) to learn domain-invariant but causal features.
    Our MAD includes an SCG that generates potential spurious correlations to diversify the source domains and an MVDC that constructs multi-view domain classifiers to remove implicit non-causal factors in latent spaces.
    Finally, MAD purifies the domain-invariant features and the causality is augmented. 
    Extensive experiments on benchmarks for cross-domain object detection verify the generalization ability to unseen domains.

\subsection*{Acknowledgments}
    \label{sec:Ack}
    This work was partially supported by National Natural Science Fund of China (62271090), National Key R\&D Program of China (2021YFB3100800), Chongqing Natural Science Fund (cstc2021jcyj-jqX0023), 
    CCF Hikvision Open Fund (CCF-HIKVISION OF 20210002), 
    CAAI-Huawei MindSpore Open Fund, 
    and Beijing Academy of Artificial Intelligence (BAAI).

{\small
\bibliographystyle{ieee_fullname}
\bibliography{egbib}
}

\end{document}